\pdfoutput=1
\documentclass{article}
\PassOptionsToPackage{table}{xcolor}
\usepackage[final]{colm2026_conference}

\usepackage{amsmath}
\usepackage{microtype}
\usepackage{hyperref}
\usepackage{url}
\usepackage{fontawesome5}
\usepackage{booktabs}
\usepackage{colortbl}
\usepackage{multirow}
\usepackage{graphicx}
\graphicspath{{figures/}}
\makeatletter
\@ifpackagelater{graphicx}{2024/06/01}{}{%
  \define@key{Gin}{alt}{}%
}
\makeatother

\usepackage{lineno}

\newcommand{\best}[1]{\textbf{\underline{#1}}}

\definecolor{darkblue}{rgb}{0, 0, 0.5}
\hypersetup{colorlinks=true, citecolor=darkblue, linkcolor=darkblue, urlcolor=darkblue}

\title{Think Before You Speak: Enabling Training-Free \\ Latent Communication Between Language Models}

\author{Rishi Khare \\
Georgia Institute of Technology\\
}

\begin{document}

\ifcolmsubmission
\linenumbers
\fi

\maketitle

\begin{abstract}
Large language models (LLMs) have demonstrated remarkable capabilities in solving complex tasks in coding, mathematical logic, and iterative reasoning, especially by coordinating effectively within multi-agent systems (MAS). However, language agents traditionally communicate via text exchange, which introduces a bottleneck in information transfer and adds prefill latency for downstream agents. Recent work proposes that directly sharing the key-value (KV) cache between language models can enable efficient latent collaboration, entirely bypassing standard text-based communication. To enable latent communication between heterogeneous models of different sizes, prior work introduces trained fusion adapters for each model pair, which limits practical deployment. We introduce CacheFuse, a training-free method which demonstrates that augmenting a downstream model with an upstream model's KV cache using linear interpolation consistently enables superior performance than the standalone downstream model can achieve. We systematically analyze CacheFuse across four reasoning and knowledge benchmarks and demonstrate up to 8.6 percentage points higher accuracy compared to single-model baselines and up to 12.2$\times$ TTFT speedup compared to text-based communication in multi-model inference. 
\end{abstract}

\vspace{-0.5em}
\begin{center}
\href{https://github.com/rishiskhare/CacheFuse}{\textcolor{black}{\large\faGithub}}\enspace
\url{https://github.com/rishiskhare/CacheFuse}
\end{center}

\section{Introduction}

Multi-agent LLM systems have become a dominant paradigm for coordinating multiple specialized models to perform complex tasks, including mathematical reasoning and software development. However, agents conventionally exchange information as natural-language text, so every inter-model exchange requires the \textit{sharer} model to autoregressively decode a message and the \textit{receiver} model to prefill the message. This text exchange dominates the time-to-first-token (TTFT) latency of downstream agents, so memory and KV cache management continue to be pervasive challenges in managing the deployment of multi-agent systems \citep{yu2026multiagentmemorycomputerarchitecture}.

A promising and practical alternative is \textit{latent} communication, in which one model transfers its internal KV cache directly to another model rather than communicating through text, allowing the receiver to attend to the sharer's cached keys and values without recomputation. Recent work demonstrates that this direct transfer can enhance task quality and speed while reducing memory consumption \citep{zou2025latentmas}. However, existing methods for transfer across \textit{heterogeneous} models, Cache-to-Cache (C2C) and Latent Cache Flow (LCF), train neural adapters that learn to project and fuse one model's cache into another model's representation space \citep{fu2026cachetocachedirectsemanticcommunication, rossi2026latentcacheflowmodeltomodel}. Because the fuser is learned uniquely per model pair, deploying the prior methods across a multi-agent communication network of $n$ models requires training up to $O(n^2)$ pair-wise adapters, which limits production serving practicality. 

Motivated by this challenge, we seek to explore practical alternatives to fuse the intelligence of larger upstream models with downstream cheaper-to-serve models in multi-model inference workloads \textit{without} training learned adapters. We observe that some model families are released across various sizes, and the models' capability generally scales with increasing model sizes \citep{kaplan2020scalinglawsneurallanguage, yang2025qwen3technicalreport}. Particular pairs of these cross-size models share identical attention geometry, with scaling in the hidden width and number of query heads. Within such an inter-model congruence, two different-sized models share an identical KV-cache shape (the same number of key/value heads, head dimension, and depth), so the sharer's per-layer cache tensors are dimensionally compatible and can be directly inserted into the receiver model. Based on this observation, we introduce CacheFuse, a training-free method to perform KV cache transfer and fusion to enable cross-model KV cache transfer between models of different size and capability. 

Our core finding is that linearly interpolating the per-layer KV caches of the sharer and receiver models improves the task performance of the downstream answering model. A linear interpolation of the receiver's own cache with a transferred upstream cache raises accuracy by up to 8.6\,pp with no training. Further, fusion can exceed both the receiver and sharer models individually. Remarkably, we find that a Qwen3-4B downstream model fused with a Qwen3-8B cache even exceeds the standalone 8B model across all of our four benchmarks. We find that downstream models in large-to-small model setups experience a higher lift in accuracy from cache fusion compared to small-to-large setups. Lastly, we demonstrate that direct cache transfer and augmentation is far more efficient than text-based message sharing between models, leading to up to a 12.2$\times$ TTFT reduction over text-to-text communication. 

\section{Related Work}
\subsection{Cross-Model KV Cache Sharing}
The development of KV cache compression and reuse frameworks is an active research area in the pursuit of memory-efficient LLM inference methods. Cache-to-Cache \citep{fu2026cachetocachedirectsemanticcommunication} introduces a pair-wise trained fuser to enable KV cache augmentation across model architectures bypassing text generation after prefilling both the sharer and receiver models. Latent Cache Flow \citep{rossi2026latentcacheflowmodeltomodel} augments learned cache-level communication by demonstrating that learned cross-model cache sharing adapters can be compressed to consume less system memory. However, these methods involve training pair-wise adapters to enable coherent KV cache fusion between agents, which limits practical deployment in serving systems. DroidSpeak \citep{liu2025droidspeakkvcachesharing} enables KV cache sharing by demonstrating that selectively profiling and recomputing \textit{critical layers} recovers quality. However, their setting focuses on sharing KV caches between downstream finetuned variants of a single base model, rather than heterogeneous transfer across models of different sizes and capability. Unlike prior methods, which involve learned pair-specific adapters for cross-model transfer, CacheFuse only requires that the two models produce KV caches that share the same geometry and combines them with a closed-form linear interpolation.

\subsection{Multi-Agent KV Cache Transfer}

Multi-agent architectures for LLM inference have become a mainstream strategy to pool the intelligence of specialized models through debate and coordination before producing a final answer. Recent work has shown that the time-to-first-token (TTFT) can be significantly reduced by sharing the raw KV cache to future agents, rather than recomputing it from scratch.
KVCOMM \citep{ye2025kvcommonlinecrosscontextkvcache} proposes KV cache reuse of previous agent outputs by aligning cache offsets for overlapping content under different prompt prefixes. CacheBlend \citep{yao2025cacheblendfastlargelanguage} and RelayCaching \citep{geng2026relaycachingacceleratingllmcollaboration} demonstrate that token profiling and selective recomputation can boost the performance of raw KV cache reuse. However, these methods have limited KV cache sharing across homogeneous models (reusing a previously computed KV cache within identical copies of the same model), whereas CacheFuse transfers caches across models of different sizes and capabilities.

\section{Method}
\subsection{Direct KV Cache Transfer}
During autoregressive generation, each decoder layer caches a key and value vector for each processed token for each KV head. For a prompt of length $S$ tokens, the generated KV cache has the shape $[n_{\mathrm{kv}}, S, d_{\mathrm{head}}]$, where $n_{\mathrm{kv}}$ is the number of key-value heads and $d_{\mathrm{head}}$ is the per-head dimension. A model with $L$ layers maintains $L$ such key and value tensors, so its KV cache shape can be defined by the dimensions $(n_{\mathrm{kv}}, d_{\mathrm{head}}, L)$. In our framing, two models enable a well-defined direct cache transfer when their KV geometries match: identical $n_{\mathrm{kv}}$, $d_{\mathrm{head}}$, and $L$ ensure that the sharer's per-layer cache has exactly the shape the receiver expects at each layer, so they can be directly inserted at the same layer position.

Our method works by sharing the KV cache of a sharer model to be used directly by a downstream receiver model, with the assumption that the sharer and receiver model share the same KV cache dimensions. The Qwen3 \citep{yang2025qwen3technicalreport} family encompasses model pairs with this key property of sharing the same KV cache geometry. Qwen3-8B and Qwen3-4B both have 36 decoder layers and share the KV cache dimensions of (8, 128, 36). Similarly, Qwen3-1.7B and Qwen3-0.6B have 28 decoder layers and share (8, 128, 28) as their KV cache dimensions. For this reason, we investigate the reuse of KV cache between these pairs, in which the number of decoder layers and KV cache geometry are identical for well-defined transfer, in both directions.

\subsection{Fusion}
\begin{figure*}[t]
  \centering
  \includegraphics[width=\textwidth,alt={CacheFuse pipeline. The sharer and receiver each prefill the same prompt, their KV caches are linearly interpolated with a scalar theta, and the receiver decodes the response from the fused cache.}]{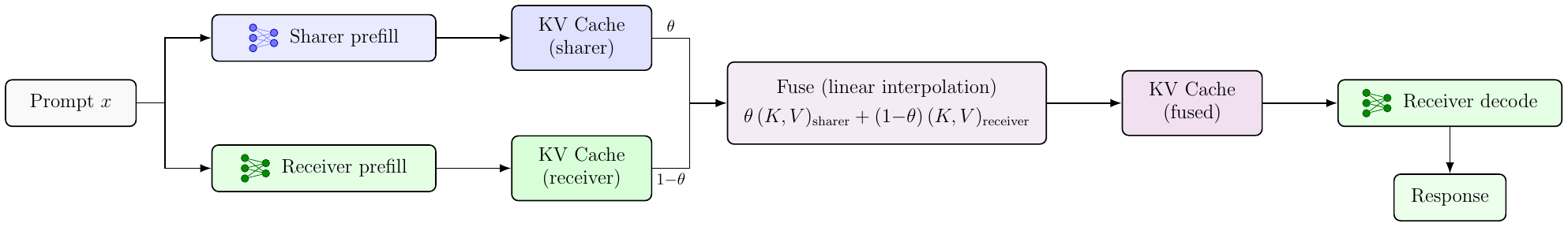}
  \caption{CacheFuse pipeline. The sharer and receiver each prefill the same prompt independently, producing per-layer KV caches of identical geometry $(n_{\mathrm{kv}}, d_{\mathrm{head}}, L)$. Every layer's key and value tensors are linearly interpolated with a single scalar $\theta$ to form a fused cache, which the downstream receiver model then decodes from directly to produce the response.}
  \label{fig:cachefuse_pipeline}
\end{figure*}

Unlike the learned fusion gates of C2C, we investigated the blending of the caches between the sharer and receiver models through a per-layer linear interpolation between KV caches as follows (shown in Equation~\eqref{eq:kv-fusion} and Figure~\ref{fig:cachefuse_pipeline}):
\begin{equation}
  (K, V)_{\mathrm{fused}} = \theta\, (K, V)_{\mathrm{sharer}} + (1-\theta)\, (K, V)_{\mathrm{receiver}} \label{eq:kv-fusion}
\end{equation}
In this setting, the single scalar $\theta \in [0,1]$ controls the proportion of the key and value tensors uniformly across layers from the sharer model that will be interpolated into the cache of the receiver model. 
At the two endpoints:
\begin{itemize}
    \item $\theta=0$: use the receiver's own cache without any transfer
    \item $\theta=1$: entire replacement with the sharer's cache
\end{itemize}

\section{Experimental Setup}

We set up multi-model benchmarks analogous to the evaluation of Cache-to-Cache and Latent Cache Flow, in which a sharer model's KV cache is directly transferred to a receiver model to decode, without any text exchange. We analyze latent cache sharing across pairs of varying sizes (0.6B, 1.7B, 4B, 8B) of the Qwen3 \citep{yang2025qwen3technicalreport} model architecture. 

\subsection{Approaches}
For each benchmark, we evaluate five approaches.
\begin{itemize}
    \item \texttt{Receiver}: the receiver model decodes normally from its own prefill ($\theta=0$, single-model baseline, no inter-model communication)
    \item \texttt{Sharer}: the sharer model decodes its own prefill (single-model baseline, no inter-model communication)
    \item \texttt{Replace}: the receiver model decodes from the transferred upstream cache entirely ($\theta=1$)
    \item \texttt{CacheFuse}: the receiver model decodes using an interpolated cache across $\theta \in (0, 1)$
    \item \texttt{Text-to-Text (T2T)}: text communication baseline in which the sharer model generates a one-sentence hint that is prepended to the receiver's prompt
\end{itemize}

Crucially, the evaluation results we report are conducted so that for each question, the receiver's cache and the sharer's (transferred) cache are each prefilled exactly once. In order to maintain fair and consistent evaluation, each sweep on the sharer model is executed by performing linear interpolation on the same prefill caches for the sharer and receiver models and only varying the single $\theta$ parameter across the $[0,1]$ range. This makes the evaluation setups controlled, as all approaches per model pair (Receiver, Sharer, Replace, CacheFuse) are scored using the exact same prefill caches to mitigate prefill variation. However, T2T is an exception to this property as it must run decode on its hint before the receiver can prefill its prompt. See Appendix~\ref{sec:appendix:prompts} for more description of the exact prompt templates we use.

\subsection{Benchmarks}
Mirroring the Latent Cache Flow evaluation suite, we analyze the accuracy of CacheFuse on four multiple-choice benchmarks: MMLU-Redux \citep{gema2025mmlu}, MMLU-Pro \citep{wang2024mmluprorobustchallengingmultitask}, ARC-Challenge \citep{clark2018thinksolvedquestionanswering}, and OpenBookQA \citep{mihaylov-etal-2018-suit}. All setups use the same generation budgets as C2C, so the text baseline is measured under the same budget the learned gating method reports. In the T2T setting, the sharer model is prompted to provide a hint which is decoded up to 256 tokens (same as C2C's T2T maximum generation limit). For the latency measurement, the answer is decoded up to 64 tokens (same as C2C's maximum response length for multiple choice). Similarly to LCF, each response is scored as an argmax of the next-token logits restricted to the option-letter tokens, so the accuracy is independent of generation limits and decode settings.

\subsection{Latency}
We adopt the same latency measurements as Latent Cache Flow: mean per-query time-to-first-token (TTFT) and time-to-end-of-answer (TTEoA). TTFT is computed after the first decoded token, so it includes all prefill, transfer, and fusion steps. All of our latency measurements are reported on a single NVIDIA A5000 GPU (24GB VRAM).

\section{Results}

We report our results as the average of two reproducible seeded runs for both accuracy and latency. Table~\ref{tab:full_sweep} shows the complete $\theta$ sweep for all benchmarks. Table~\ref{tab:arc_obqa_accuracy_comparison} summarizes accuracy for the best $\theta$ per pair alongside baselines. Table~\ref{tab:latency_all_pairs} reports latency (TTFT and TTEoA).

\begin{table*}[t]
  \centering\scriptsize
  \setlength{\tabcolsep}{3pt}
  \renewcommand{\arraystretch}{1.3}
  \begin{tabular}{ll *{11}{r} | rr}
    \toprule
    \textbf{Pair} & \textbf{Benchmark}
      & \textbf{Receiver} & \textbf{0.1} & \textbf{0.2} & \textbf{0.3} & \textbf{0.4}
      & \textbf{0.5} & \textbf{0.6} & \textbf{0.7} & \textbf{0.8} & \textbf{0.9}
      & \textbf{Replace} & \textbf{Sharer} & \textbf{T2T} \\
    \midrule
    \multirow{4}{*}{8B $\to$ 4B}
      & ARC-C  & 87.03 & 87.97 & 88.82 & 89.33 & 90.70 & 91.47 & \best{91.72} & 91.21 & 90.96 & 90.02 & 83.45 & 91.21 & 90.19 \\
      & OBQA   & 78.00 & 79.00 & 80.80 & 83.40 & 84.40 & \best{86.60} & 85.20 & 84.00 & 83.00 & 82.60 & 71.40 & 85.00 & 83.80 \\
      & MMLU-R & 69.95 & 71.18 & 72.40 & 73.28 & 74.09 & \best{74.70} & 74.68 & 73.98 & 73.42 & 72.18 & 63.11 & 73.88 & 72.26 \\
      & MMLU-P & 41.08 & 42.20 & 43.46 & 44.33 & 45.07 & \best{45.33} & 44.93 & 44.22 & 42.80 & 37.43 & 27.81 & 44.41 & 43.87 \\
    \midrule
    \multirow{4}{*}{4B $\to$ 8B}
      & ARC-C  & 91.21 & 91.30 & 91.30 & 91.64 & 91.55 & \best{91.72} & 90.61 & 89.76 & 89.51 & 77.30 & 39.16 & 87.03 & 91.98 \\
      & OBQA   & 85.00 & 85.60 & 85.00 & \best{86.00} & 85.80 & 85.80 & 84.80 & 82.60 & 84.00 & 78.40 & 54.60 & 78.00 & 87.80 \\
      & MMLU-R & 73.88 & 74.32 & 74.40 & 74.53 & 74.49 & \best{74.77} & 73.81 & 72.60 & 71.88 & 59.26 & 35.60 & 69.95 & 75.60 \\
      & MMLU-P & 44.41 & \best{44.79} & 44.76 & 44.27 & 43.24 & 41.41 & 41.43 & 38.65 & 36.85 & 20.16 & 12.81 & 41.08 & 46.85 \\
    \midrule
    \multirow{4}{*}{1.7B $\to$ 0.6B}
      & ARC-C  & 52.65 & 56.23 & \best{61.09} & 60.67 & 52.47 & 38.99 & 29.95 & 26.54 & 25.60 & 26.45 & 24.40 & 73.89 & 59.39 \\
      & OBQA   & 53.60 & 54.20 & \best{54.60} & 53.40 & 47.60 & 36.80 & 32.20 & 26.80 & 25.00 & 23.60 & 24.60 & 65.40 & 52.40 \\
      & MMLU-R & 43.26 & 44.98 & \best{47.93} & 46.77 & 40.00 & 33.28 & 29.37 & 26.88 & 25.16 & 24.86 & 24.60 & 57.42 & 46.54 \\
      & MMLU-P & 18.81 & 20.10 & \best{20.96} & 20.30 & 15.77 & 13.79 & 12.68 & 12.23 & 11.54 & 11.74 & 11.22 & 29.13 & 19.73 \\
    \midrule
    \multirow{4}{*}{0.6B $\to$ 1.7B}
      & ARC-C  & 73.89 & \best{74.74} & 73.63 & 72.01 & 69.80 & 33.79 & 23.38 & 22.70 & 23.89 & 23.63 & 23.72 & 52.65 & 70.31 \\
      & OBQA   & 65.40 & \best{66.20} & 65.20 & 59.60 & 62.20 & 35.00 & 27.80 & 29.00 & 26.60 & 25.00 & 25.00 & 53.60 & 61.60 \\
      & MMLU-R & \best{57.42} & 57.05 & 56.12 & 55.32 & 52.95 & 28.98 & 24.00 & 23.74 & 22.95 & 23.91 & 25.02 & 43.26 & 52.11 \\
      & MMLU-P & 29.13 & \best{29.56} & 29.34 & 25.58 & 20.98 & 12.28 & 11.54 & 11.87 & 11.20 & 11.20 & 11.38 & 18.81 & 26.25 \\
    \bottomrule
  \end{tabular}
  \caption{Full $\theta$ sweep across benchmarks on Qwen3 model pairs. \textbf{Receiver} corresponds to $\theta = 0$ (pure receiver cache). Columns 0.1–0.9 are $\theta$ values for the CacheFuse linear interpolation. \textbf{Replace} corresponds to $\theta = 1$ (pure sharer cache). \textbf{Sharer} denotes the standalone sharer model (reference). \textbf{T2T} denotes the text-to-text transfer baseline (reference). Bold underlined marks the best score in the $\theta$ sweep (across the Receiver to Replace endpoints).}
  \label{tab:full_sweep}
\end{table*}

\begin{table*}[t]
  \centering
  \scriptsize
  \setlength{\tabcolsep}{9pt}
  \renewcommand{\arraystretch}{1.2}
  \begin{tabular}{llrrrr}
    \toprule
    \textbf{Pair} & \textbf{Method} & \textbf{ARC-C} & \textbf{OBQA} & \textbf{MMLU-R} & \textbf{MMLU-P} \\
    \midrule
    \multirow{4}{*}{Qwen3-8B $\rightarrow$ Qwen3-4B}
      & Sharer                    & 91.21 & 85.00 & 73.88 & 44.41 \\
      & Receiver                  & 87.03 & 78.00 & 69.95 & 41.08 \\
      & \cellcolor{violet!10}CacheFuse    & \cellcolor{violet!10}\best{91.72} & \cellcolor{violet!10}\best{86.60} & \cellcolor{violet!10}\best{74.70} & \cellcolor{violet!10}\best{45.33} \\
      & T2T                       & 90.19 & 83.80 & 72.26 & 43.87 \\
    \midrule
    \multirow{4}{*}{Qwen3-4B $\rightarrow$ Qwen3-8B}
      & Sharer                    & 87.03 & 78.00 & 69.95 & 41.08 \\
      & Receiver                  & 91.21 & 85.00 & 73.88 & 44.41 \\
      & \cellcolor{violet!10}CacheFuse    & \cellcolor{violet!10}\best{91.72} & \cellcolor{violet!10}\best{86.00} & \cellcolor{violet!10}\best{74.77} & \cellcolor{violet!10}\best{44.79} \\
      & T2T                       & 91.98 & 87.80 & 75.60 & 46.85 \\
    \midrule
    \multirow{4}{*}{Qwen3-1.7B $\rightarrow$ Qwen3-0.6B}
      & Sharer                    & 73.89 & 65.40 & 57.42 & 29.13 \\
      & Receiver                  & 52.65 & 53.60 & 43.26 & 18.81 \\
      & \cellcolor{violet!10}CacheFuse    & \cellcolor{violet!10}\best{61.09} & \cellcolor{violet!10}\best{54.60} & \cellcolor{violet!10}\best{47.93} & \cellcolor{violet!10}\best{20.96} \\
      & T2T                       & 59.39 & 52.40 & 46.54 & 19.73 \\
    \midrule
    \multirow{4}{*}{Qwen3-0.6B $\rightarrow$ Qwen3-1.7B}
      & Sharer                    & 52.65 & 53.60 & 43.26 & 18.81 \\
      & Receiver                  & 73.89 & 65.40 & \best{57.42} & 29.13 \\
      & \cellcolor{violet!10}CacheFuse    & \cellcolor{violet!10}\best{74.74} & \cellcolor{violet!10}\best{66.20} & \cellcolor{violet!10}57.42 & \cellcolor{violet!10}\best{29.56} \\
      & T2T                       & 70.31 & 61.60 & 52.11 & 26.25 \\
    \bottomrule
  \end{tabular}
  \caption{Accuracy across all four benchmarks for the best-performing $\theta$ per pair (exact $\theta$ per benchmark given in Table~\ref{tab:full_sweep}). Bold underlined marks where CacheFuse exceeds Receiver (otherwise Receiver is marked).}
  \label{tab:arc_obqa_accuracy_comparison}
\end{table*}

\begin{table*}[t]
  \centering
  \scriptsize
  \setlength{\tabcolsep}{7pt}
  \renewcommand{\arraystretch}{1.2}
  \begin{tabular}{llrrrr}
    \toprule
    \textbf{Pair} & \textbf{Method} &
    \multicolumn{2}{c}{\textbf{ARC-C}} &
    \multicolumn{2}{c}{\textbf{OBQA}} \\
    \cmidrule(lr){3-4}\cmidrule(lr){5-6}
    & & \textbf{TTFT} & \textbf{TTEoA} & \textbf{TTFT} & \textbf{TTEoA} \\
    \midrule
    \multirow{5}{*}{Qwen3-8B $\rightarrow$ Qwen3-4B}
      & Receiver &   78.8 & 2518.8 &   78.0 & 2497.2 \\
      & Replace  &   79.5 & 2518.3 &   78.6 & 2477.1 \\
      & Sharer   &   80.2 & 2569.0 &   79.3 & 2545.7 \\
      & \cellcolor{violet!10}CacheFuse     &  \cellcolor{violet!10}120.5 & \cellcolor{violet!10}2560.3 & \cellcolor{violet!10}119.0 & \cellcolor{violet!10}2536.8 \\
      & T2T      & 1468.5 & 3915.6 & 1259.1 & 3680.0 \\
    \midrule
    \multirow{5}{*}{Qwen3-4B $\rightarrow$ Qwen3-8B}
      & Receiver &   79.7 & 2563.1 &  80.2 & 2587.6 \\
      & Replace  &   79.9 & 2563.8 &  80.4 & 2588.0 \\
      & Sharer   &   79.0 & 2512.6 &  79.5 & 2534.9 \\
      & \cellcolor{violet!10}CacheFuse     &  \cellcolor{violet!10}121.0 & \cellcolor{violet!10}2605.7 & \cellcolor{violet!10}121.7 & \cellcolor{violet!10}2629.2 \\
      & T2T      &  952.0 & 3439.2 & 813.4 & 3318.5 \\
    \midrule
    \multirow{5}{*}{Qwen3-1.7B $\rightarrow$ Qwen3-0.6B}
      & Receiver &  60.5 & 1899.9 &  60.1 & 1884.8 \\
      & Replace  &  60.6 & 1885.1 &  60.0 & 1881.7 \\
      & Sharer   &  61.6 & 1968.5 &  61.0 & 1954.5 \\
      & \cellcolor{violet!10}CacheFuse     &  \cellcolor{violet!10}92.5 & \cellcolor{violet!10}1932.7 & \cellcolor{violet!10}91.7 & \cellcolor{violet!10}1916.5 \\
      & T2T      & 794.9 & 2637.9 & 710.4 & 2535.7 \\
    \midrule
    \multirow{5}{*}{Qwen3-0.6B $\rightarrow$ Qwen3-1.7B}
      & Receiver &  61.0 & 1951.6 &  61.1 & 1957.2 \\
      & Replace  &  61.1 & 1951.7 &  61.5 & 1957.8 \\
      & Sharer   &  60.0 & 1884.1 &  60.4 & 1890.2 \\
      & \cellcolor{violet!10}CacheFuse     &  \cellcolor{violet!10}92.7 & \cellcolor{violet!10}1984.0 & \cellcolor{violet!10}93.0 & \cellcolor{violet!10}1989.5 \\
      & T2T      & 384.2 & 2276.2 & 341.6 & 2237.5 \\
    \bottomrule
  \end{tabular}
  \caption{Latency in milliseconds (ms) across model pairs and transfer methods. TTFT denotes time-to-first-token. TTEoA denotes time-to-end-of-answer.}
  \label{tab:latency_all_pairs}
\end{table*}

\subsection{Fusion Can Exceed Both Models}
\begin{figure}[t]
  \centering
  \includegraphics[alt={MMLU-R accuracy versus theta for Qwen3-8B sharing into Qwen3-4B. CacheFuse peaks near theta 0.5 where it beats both the receiver and the larger sharer.}]{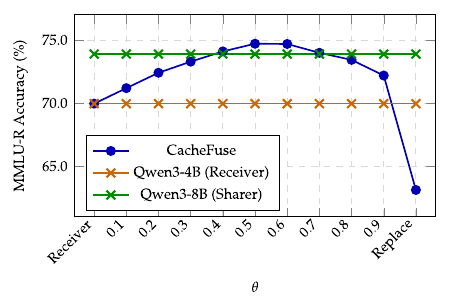}
  \caption{MMLU-R accuracy sweep for Qwen3-8B $\rightarrow$ Qwen3-4B. Receiver ($\theta = 0$) and Replace ($\theta = 1$) are included as endpoints. Fusion peaks at $\theta = 0.5$, exceeding the receiver model individually (Qwen3-4B), direct KV cache replacement, and even the larger sharer model individually (Qwen3-8B).}
  \label{fig:fusion_sweep}
\end{figure}

Our primary result is that a linear combination of both KV caches can beat both models individually and consistently outperforms the receiver model alone. Table~\ref{tab:full_sweep} gives the full $\theta$ sweep results across all benchmarks and pairs, compared against T2T, Sharer, and Receiver. Table~\ref{tab:arc_obqa_accuracy_comparison} summarizes best-$\theta$ CacheFuse accuracy against Sharer, Receiver, and T2T. For Qwen3-8B $\rightarrow$ Qwen3-4B, the fused model ($\theta=0.5$ to $0.6$) consistently exceeds the receiver model alone on all benchmarks by up to 8.6\,pp on OBQA. Although in this specific setting, the fused variant is run on Qwen3-4B, we find that this even exceeds the larger 8B sharer model by up to 1.6\,pp on OBQA. Figure~\ref{fig:fusion_sweep} illustrates this for Qwen3-8B $\rightarrow$ Qwen3-4B on MMLU-Redux, where fusion peaks above both the receiver and sharer baselines. We further provide results from experiments with selecting $\theta$ via an offline calibration sweep on held-out examples in Appendix~\ref{sec:appendix:calibration}.

\subsection{Large-to-Small vs. Small-to-Large Cache Transfer} 
We also find that the utility of cache transfer strongly depends on which model shares its cache and which model receives it to produce an answer. When a \textit{stronger} model shares its cache with a downstream \textit{weaker} model (i.e. 1.7B $\rightarrow$ 0.6B and 8B $\rightarrow$ 4B scenarios), fusion yields larger gains over the receiver alone than weak-to-strong model transfers. When a \textit{stronger} model consumes a \textit{weaker} model's cache (0.6B $\rightarrow$ 1.7B and 4B $\rightarrow$ 8B scenarios), we observe a comparatively incremental benefit against the receiver alone (at most 1\,pp accuracy improvement). This aligns with the expectation that in a small-to-large transfer, the stronger downstream receiver model has minimal upside from a weak sharer model.

\subsection{Replacement Compatibility}
Direct cache replacement ($\theta=1$) serves as an informative signal on raw cache compatibility. In every pair apart from 8B $\rightarrow$ 4B, direct cache replacement approximately collapses to the regime of random chance correctness in multiple-choice question-answering, which indicates that a raw cross-model cache is generally not directly usable. However, for the 8B $\rightarrow$ 4B pair, the raw 8B cache is highly usable on its own (the replacement strategy achieves about 83.5\% on ARC-C), though the reverse direction (4B $\rightarrow$ 8B) experiences much lower coherence (39.2\% on ARC-C).

\subsection{Latency}
\begin{figure}[t]
  \begin{minipage}[t]{0.48\textwidth}
    \centering
    \includegraphics[alt={Bar chart of time-to-first-token speedup over text-to-text communication for Qwen3-8B to Qwen3-4B on ARC-Challenge. CacheFuse is 12.2 times faster than text-to-text.}]{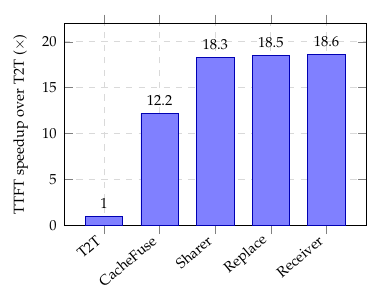}
    \caption{TTFT speedup over T2T for Qwen3-8B $\rightarrow$ Qwen3-4B on ARC-C. Linearly interpolating the sharer and reciever caches (CacheFuse) achieves $12.2\times$ faster TTFT than dense prefill of text-to-text communication (T2T). Single-model baselines are ${\sim}18\times$ faster.}
    \label{fig:ttft_bar}
  \end{minipage}%
  \hfill
  \begin{minipage}[t]{0.48\textwidth}
    \centering
    \includegraphics[alt={Time-to-first-token versus ARC-Challenge accuracy for Qwen3-8B to Qwen3-4B. CacheFuse has the highest accuracy at much lower latency than text-to-text.}]{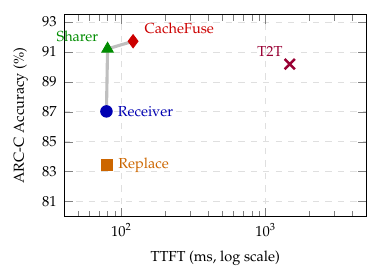}
    \caption{TTFT--accuracy Pareto plot for Qwen3-8B $\rightarrow$ Qwen3-4B on ARC-C. Lower TTFT and higher accuracy (top left) is better. The gray line traces the Pareto frontier. CacheFuse achieves the highest accuracy at significantly lower TTFT than T2T.}
    \label{fig:pareto_8b_4b}
  \end{minipage}
\end{figure}

Table~\ref{tab:latency_all_pairs} shows that the latent methods are substantially more computationally efficient than text communication. The text-to-text (T2T) baseline is far slower because the sharer must autoregressively generate its hint message to share to the receiver (up to 256-token maximum communication tokens). We observe that TTFT, especially for T2T, scales with the sharer model's size from 384 ms (0.6B as the sharer model) to 1469 ms (8B as the sharer model) on the text-to-text baselines. TTEoA is dominated by the answer decode generation (up to 64-token maximum limit) and experiences a smaller factor of speedup between the approaches compared to TTFT. Figure~\ref{fig:ttft_bar} shows the TTFT breakdown for the Qwen3-8B $\rightarrow$ Qwen3-4B pair on ARC-C. Figure~\ref{fig:pareto_8b_4b} displays a TTFT--accuracy Pareto plot for the same setting.

\section{Conclusion}
CacheFuse demonstrates that answering from another model's latent representation instead of text can be accomplished across models of different sizes without a trained adapter, given the setting involves sharing the cache between models with congruent KV cache geometries. Our findings demonstrate that across both large-to-small model and small-to-large model KV cache transfers, linear interpolation between the models' KV caches (i) improves the performance of the downstream model, (ii) can exceed both models in the pair, and (iii) substantially reduces TTFT over text-based communication.

\section*{Limitations}
In order to perform direct KV cache transfer, CacheFuse requires the sharer and receiver caches to share the same KV cache dimensions for well-defined reuse. In addition, our experimental setup analyzes model pairs that share the same tokenizer (Qwen3 in this work). An exciting direction for future work is the development of methods for cross-model KV cache transfer across entirely heterogeneous tokenizer and model architecture families (i.e. Qwen to Llama).

\section*{Acknowledgements} I thank Zishen Wan for his guidance and helpful discussions during the development of this work.

\bibliography{colm2026_conference}
\bibliographystyle{colm2026_conference}

\appendix
\section{Appendix}
\label{sec:appendix}

\subsection*{Prompts}
\label{sec:appendix:prompts}

\begin{samepage}
Every evaluated approach scores responses using the following zero-shot prompt, where next-token logits on \texttt{A}/\texttt{B}/\texttt{C}/\texttt{D} (up to 10 choices for MMLU-Pro) determine accuracy:

\noindent\includegraphics[width=\textwidth,alt={Evaluation prompt template. The model is given a multiple-choice question and options and is asked to answer.}]{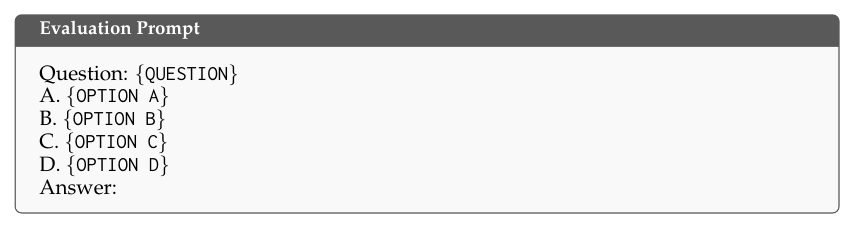}
\end{samepage}

\begin{samepage}
For the text-to-text (T2T) baseline, we follow the same sharer instruction from Cache-to-Cache~\citep{fu2026cachetocachedirectsemanticcommunication}. The sharer model is prompted:

\noindent\includegraphics[width=\textwidth,alt={Sharer hint prompt. The sharer is asked for one sentence of essential background and told not to solve the question or give the answer.}]{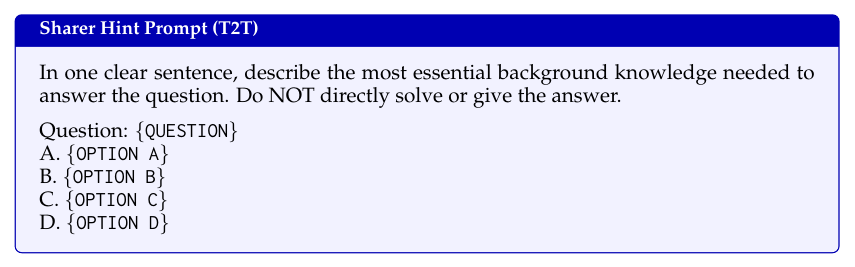}
\end{samepage}

\noindent\begin{minipage}{\textwidth}
\indent The receiver model then receives the sharer's output as a hint:

\vspace{0.45em}
\includegraphics[width=\linewidth,alt={Receiver prompt with hint. The sharer's sentence is prepended as background, followed by the multiple-choice question.}]{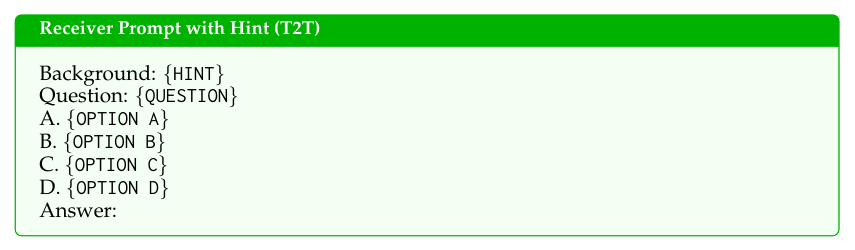}
\end{minipage}

We design our prompt templates to be minimal and lean compared to the additional instructions provided within C2C's prompt templates to faithfully reflect the task quality of the language models without the influence of prompt engineering.

\subsection*{Offline Theta Calibration}
\label{sec:appendix:calibration}

\begin{table*}[t]
  \centering\scriptsize
  \setlength{\tabcolsep}{3.5pt}
  \renewcommand{\arraystretch}{1.3}
  \begin{tabular}{ll *{11}{r} | r}
    \toprule
    \textbf{Pair} & \textbf{Benchmark}
      & \textbf{Receiver} & \textbf{0.1} & \textbf{0.2} & \textbf{0.3} & \textbf{0.4}
      & \textbf{0.5} & \textbf{0.6} & \textbf{0.7} & \textbf{0.8} & \textbf{0.9}
      & \textbf{Replace} & \textbf{Sharer} \\
    \midrule
    \multirow{2}{*}{Qwen3-8B $\to$ Qwen3-4B}
      & ARC-C  & 87.8 & 88.8 & 89.7 & 90.0 & 91.2 & 91.6 & 91.5 & \best{92.0} & 91.9 & 90.9 & 83.6 & 92.0 \\
      & OBQA   & 74.5 & 75.9 & 77.3 & 78.5 & \best{79.8} & 79.4 & 79.3 & 79.6 & 78.8 & 77.6 & 70.1 & 79.5 \\
    \midrule
    \multirow{2}{*}{Qwen3-4B $\to$ Qwen3-8B}
      & ARC-C  & 92.0 & 92.1 & 92.3 & \best{92.4} & 92.1 & 91.7 & 91.8 & 90.5 & 89.4 & 77.6 & 41.4 & 87.8 \\
      & OBQA   & 79.5 & 80.1 & 80.0 & \best{80.3} & 79.9 & \best{80.3} & 79.6 & 78.5 & 77.3 & 71.6 & 57.4 & 74.5 \\
    \midrule
    \multirow{2}{*}{Qwen3-1.7B $\to$ Qwen3-0.6B}
      & ARC-C  & 57.5 & 61.6 & \best{65.7} & 64.9 & 51.4 & 38.4 & 30.3 & 26.0 & 23.5 & 21.8 & 23.2 & 75.2 \\
      & OBQA   & 46.7 & 49.0 & 50.8 & \best{53.7} & 50.9 & 43.1 & 34.7 & 31.5 & 28.3 & 25.5 & 26.6 & 60.8 \\
    \midrule
    \multirow{2}{*}{Qwen3-0.6B $\to$ Qwen3-1.7B}
      & ARC-C  & 75.2 & 75.1 & \best{76.8} & 74.4 & 71.4 & 34.0 & 22.3 & 22.9 & 22.6 & 24.0 & 21.5 & 57.5 \\
      & OBQA   & 60.8 & 62.1 & \best{62.4} & 60.4 & 58.6 & 42.6 & 36.1 & 35.4 & 36.5 & 33.8 & 28.1 & 46.7 \\
    \bottomrule
  \end{tabular}
  \caption{Accuracy of offline $\theta$ calibration sweep on the first 1000 examples of each \texttt{train} split (ARC-C and OBQA), averaged over seeds. Column conventions match Table~\ref{tab:full_sweep}: \textbf{Receiver} corresponds to $\theta = 0$. Columns 0.1–0.9 are linear interpolation between the sharer and receiver. \textbf{Replace} corresponds to $\theta = 1$. \textbf{Sharer} denotes the standalone sharer. Bold underlined marks the best score in the $\theta$ sweep (across Receiver to Replace endpoints).}
  \label{tab:train_calib_sweep}
\end{table*}

\begin{figure*}[t]
  \centering
  \includegraphics[width=\textwidth,alt={Calibration versus test theta sweeps on ARC-Challenge. Left, Qwen3-8B to Qwen3-4B. Right, Qwen3-1.7B to Qwen3-0.6B.}]{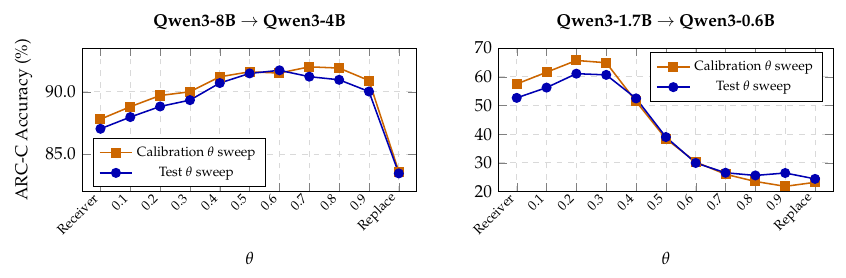}
  \caption{Calibration vs.\ test $\theta$ sweeps on ARC-C. Left: Qwen3-8B $\rightarrow$ Qwen3-4B (calibration peaks at $\theta = 0.7$ while test peaks at $\theta = 0.6$). Right: Qwen3-1.7B $\rightarrow$ Qwen3-0.6B (both peak at $\theta = 0.2$).}
  \label{fig:calib_vs_test_sweep}
\end{figure*}

\begin{table*}[t]
  \centering\scriptsize
  \setlength{\tabcolsep}{3.5pt}
  \renewcommand{\arraystretch}{1.25}
  \begin{tabular}{llccccc}
    \toprule
    \textbf{Pair} & \textbf{Benchmark}
      & \textbf{Receiver} & \textbf{Test peak}
      & \textbf{Top-1} & \textbf{Top-2} & \textbf{Top-3} \\
    \midrule
    \multirow{2}{*}{Qwen3-8B$\to$Qwen3-4B}
      & ARC-C  & 87.03 & 91.72 ($\theta{=}0.6$) & 91.21 ($\theta{=}0.7$) & 91.21 ($\theta{=}0.7$) & 91.47 ($\theta{=}0.5$) \\
      & OBQA   & 78.00 & 86.60 ($\theta{=}0.5$) & 84.40 ($\theta{=}0.4$) & 84.40 ($\theta{=}0.4$) & \best{86.60} ($\theta{=}0.5$) \\
    \midrule
    \multirow{2}{*}{Qwen3-4B$\to$Qwen3-8B}
      & ARC-C  & 91.21 & 91.72 ($\theta{=}0.5$) & 91.64 ($\theta{=}0.3$) & 91.64 ($\theta{=}0.3$) & 91.64 ($\theta{=}0.3$) \\
      & OBQA   & 85.00 & 86.00 ($\theta{=}0.3$) & \best{86.00} ($\theta{=}0.3$) & \best{86.00} ($\theta{=}0.3$) & \best{86.00} ($\theta{=}0.3$) \\
    \midrule
    \multirow{2}{*}{Qwen3-1.7B$\to$Qwen3-0.6B}
      & ARC-C  & 52.65 & 61.09 ($\theta{=}0.2$) & \best{61.09} ($\theta{=}0.2$) & \best{61.09} ($\theta{=}0.2$) & \best{61.09} ($\theta{=}0.2$) \\
      & OBQA   & 53.60 & 54.60 ($\theta{=}0.2$) & 53.40 ($\theta{=}0.3$) & 53.40 ($\theta{=}0.3$) & \best{54.60} ($\theta{=}0.2$) \\
    \midrule
    \multirow{2}{*}{Qwen3-0.6B$\to$Qwen3-1.7B}
      & ARC-C  & 73.89 & 74.74 ($\theta{=}0.1$) & 73.63 ($\theta{=}0.2$) & 73.89 ($\theta{=}0.0$) & \best{74.74} ($\theta{=}0.1$) \\
      & OBQA   & 65.40 & 66.20 ($\theta{=}0.1$) & 65.20 ($\theta{=}0.2$) & \best{66.20} ($\theta{=}0.1$) & \best{66.20} ($\theta{=}0.1$) \\
    \bottomrule
  \end{tabular}
  \caption{Top-$k$ offline calibration for reducing the test-time $\theta$ search. \textbf{Test peak} is the best accuracy (\%) over the full test $\theta$ sweep (Table~\ref{tab:full_sweep}). \textbf{Top-$k$} selects the $k$ highest-accuracy $\theta$ values from the averaged 1000-example \texttt{train} calibration sweep (Table~\ref{tab:train_calib_sweep}), then reports the best test accuracy among only those $k$ candidates (selected $\theta$ in parentheses). Bold underlined denotes where the top-$k$ sweep exactly recovers the oracle score found on the test time search.}
  \label{tab:calib_transfer}
\end{table*}

We demonstrate that an effective $\theta$ parameter can be approximated via an offline calibration sweep, rather than a full $\theta$ sweep at test time. On two of the benchmarks that we evaluate on (ARC-Challenge and OpenBookQA), the authors provide a \texttt{train} split. In Table~\ref{tab:train_calib_sweep}, we report the full $\theta$ sweep averaged over the first 1000 examples of each \texttt{train} split across both seeds, which can be compared against the corresponding \texttt{test} split results in Table~\ref{tab:full_sweep}. Figure~\ref{fig:calib_vs_test_sweep} overlays the calibration and test $\theta$ sweeps on ARC-C for the 8B $\rightarrow$ 4B and 1.7B $\rightarrow$ 0.6B pairs.

We suggest that the search space for $\theta$ at test time can be greatly reduced by selecting the top-$k$ best-performing $\theta$ values during an initial calibration pass of representative held out examples. By limiting our calibration to an equal-sized limited subset of 1000 held out examples per benchmark, we demonstrate in Table~\ref{tab:calib_transfer} that selecting between the top 3 best $\theta$ values found to give the highest accuracies in the 1000-example \texttt{train} subset consistently outperforms the receiver alone and achieves near-optimal performance to test-time calibration when applied to the \texttt{test} set. 

\subsection*{Runtime Settings}
For reproducibility, we perform all reported experiments with seed values of 40 and 42 on CUDA 13.0 and transformers 5.12.1 and report the average of the two runs for both latency and accuracy measurements.

\end{document}